\documentclass[runningheads]{llncs}
\usepackage{graphicx}

\usepackage{pgfplots}
\usepackage{tikz}
\usepackage{mathrsfs}
\usepackage{amsfonts}
\usepackage[ruled]{algorithm2e}
\usepackage{floatrow}
\newfloatcommand{capbtabbox}{table}[][\FBwidth]

\SetCommentSty{mycommfont}

\begin{document}
	%
	%\title{Contribution Title\thanks{Supported by organization x.}}
	\title{Cross-Modality Multi-Atlas Segmentation Using Deep Neural Networks}
	
	\author{Wangbin Ding \inst{1} \and
		Lei Li\inst{2,3,4} \and
		Xiahai Zhuang\inst{2,*} \and
		Liqin Huang\inst{1,*}}
	
	% index{Ding, Wangbin; Li, Lei; Huang, Liqin; Zhuang, Xiahai} 
	
	\authorrunning{Ding et al.}
	\institute{College of Physics and Information Engineering, Fuzhou University, Fuzhou, China 
		\and School of Data Science, Fudan University, Shanghai, China
		\and School of Biomedical Engineering, Shanghai Jiao Tong University, Shanghai, China  
		\and School of Biomedical Engineering and Imaging Sciences, King’s College London, London, UK
	}
	\footnotetext{* X Zhuang and L Huang are co-senior and corresponding authors: zxh@fudan.edu.cn; hlq@fzu.edu.cn. This work was funded by the National Natural Science Foundation of China (Grant No. 61971142), and Shanghai Municipal Science and Technology Major Project (Grant No. 2017SHZDZX01).}
	%	ABC Institute, Rupert-Karls-University Heidelberg, Heidelberg, Germany\\
	%	\email{\{abc,lncs\}@uni-heidelberg.de}
	%	}
	%
	\maketitle              % typeset the header of the contribution
	\begin{abstract}
		Both image registration and label fusion in the multi-atlas segmentation (MAS) rely on the intensity similarity between target and atlas images. However, such similarity can be problematic when target and atlas images are acquired using different imaging protocols. High-level structure information can provide reliable similarity measurement for cross-modality images when cooperating with deep neural networks (DNNs). This work presents a new MAS framework for cross-modality images, where both image registration and label fusion are achieved by DNNs. For image registration, we propose a consistent registration network, which can jointly estimate forward and backward dense displacement fields (DDFs). Additionally, an invertible constraint is employed in the network to reduce the correspondence ambiguity of the estimated DDFs. For label fusion, we adapt a few-shot learning network to measure the similarity of atlas and target patches. Moreover, the network can be seamlessly integrated into the patch-based label fusion. The proposed framework is evaluated on the MM-WHS dataset of MICCAI 2017. Results show that the framework is effective in both cross-modality registration and segmentation.
		
		\keywords{MAS  \and Cross-Modality  \and Similarity.}
	\end{abstract}
	
	\section{Introduction}
	Segmentation is an essential step for medical image processing. Many clinical applications rely on an accurate segmentation to extract specific anatomy or compute some functional indices. The multi-atlas segmentation (MAS) has proved to be an effective method for medical image segmentation \cite{ref_article0}. Generally, it  contains two steps, i.e., a pair-wise registration between target image and atlases, and a label fusion among selected reliable atlases. Conventional MAS methods normally process images from single modality,  but in many scenarios they could benefit from cross-modality image processing \cite{ref_article1}. To obtain such a method, registration and label fusion algorithms that can adapt to cross-modality data are required.
	\begin{figure}[tb] 
		\centering
		\includegraphics[width=0.94\textwidth]{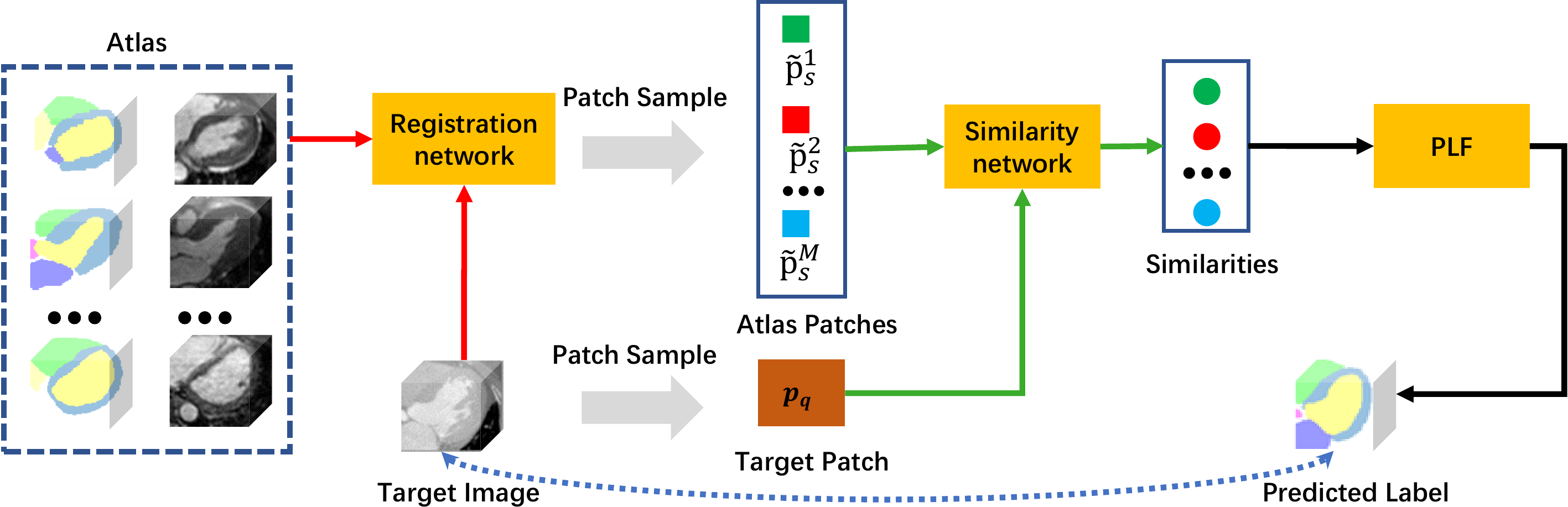}
		\caption{The pipeline of the cross-modality MAS framework. The atlases are first warped to the target image by the registration model (see section \ref{sec:reg}). Thus, the warped atlas label becomes a candidate segmentation of the target image simultaneously. Then, each voting patch sampled from warped atlases is weighted according to its similarity to the corresponding target patch (see section \ref{sec:sim}). Based on the weight, one can obtain a final label for the target image using the PLF strategy.}
		\label{fig:1}    
	\end{figure}
	
	To achieve cross-modality registration, a common approach is to design a modality-invariance similarity as the registration criterion, such as mutual information (MI) \cite{ref_article2}, normalized mutual information (NMI) \cite{ref_article3}. An alternative way is to employ structural representations of images, which are supposed to be invariant across multi-modality images  \cite{ref_article4,ref_article5}. Recently, several deep learning (DL) based multi-modality registration algorithms are developed. For example, Hu et al. proposed a weakly-supervised multi-modality registration network by exploring the dense voxel correspondence from anatomical labels \cite{ref_article6}. Qin et al. designed an unsupervised registration network based on disentangled shape representations, and then converted the multi-modality registration into a mono-modality problem in the latent shape space  \cite{ref_article7}.
	
	For label fusion, there are several widely utilized strategies, such as majority voting (MV), plurality voting, global or local weighted voting, joint label fusion (JLF) \cite{ref_article8}, statistical modeling approach \cite{ref_article9}, and patch-based label fusion (PLF)  \cite{ref_article18}. To use cross-modality atlas, Kasiri et al. presented a similarity measurement based on un-decimated wavelet transform for cross-modality atlas fusion \cite{kasiri2014cross}. Furthermore, Zhuang et al. proposed a multi-scale patch strategy to extract multi-level structural information for multi-modality atlas fusion \cite{ref_article10}. Recently, learning methods are engaged to improve the performance of label fusion. Ding et al. proposed a DL-based label fusion strategy, namely VoteNet, which can locally select reliable atlases and fuse atlas labels by plurality voting \cite{ref_article11}. To enhance PLF strategy, Sanroma et al. and Yang et al. attempted to achieve a better deep feature similarity between target and atlas patches through deep neural networks (DNN)  \cite{ref_article12,ref_article13}. Similarly, Xie et al. incorporated a DNN to predict the weight of voting patches for the JLF strategy \cite{ref_article14}. All these learning-based label fusion works assumed that atlas and target images come from the same modality.

	This work is aimed at designing a DNN-based approach to achieve accurate registration and label fusion for cross-modality MAS. Figure \ref{fig:1} presents the pipeline of our proposed MAS method. The main contributions of this work are summarized as follows:
	(1) We present a DNN-based MAS framework for cross-modality segmentation, and validate it using the MM-WHS dataset \cite{ref_article0}.
	(2) We propose a consistent registration network, where an invertible constraint is employed to encourage the uniqueness of transformation fields between cross-modality images. 
	(3) We introduce a similarity network based on few-shot learning, which can estimate the patch-based similarity between the target and atlas images.
	
	\section{Method}
	\subsection{Consistent Registration Network}
	\label{sec:reg}
	%	In this section, we first provide the general architecture of the registration network. Then we explain the benefit of consistent constraint.
	\subsubsection{Network Architecture:}
	Suppose given $N$ atlases $\{ (I_a^1,L_a^1 ),\cdots,(I_a^N,L_a^N) \}$ and a target ($I_t$,$L_t$), for each pair of $I_a^i$ and $I_t$, two registration procedures could be performed by switching the role of $I_a^i$ and $I_t$. We denote the dense displacement field (DDF) from $I_a^i$ to $I_t$ as $U^i$, and vice versa as $V^i$. For convenience, we abbreviate $I_a^i$, $L_a^i$, $U^i$ and $V^i$ as $I_a$, $L_a$, $U$ and $V$  when no confusion is caused. Consider the label as a mapping function from common spatial space to label space: $\Omega \to \mathbb{L}$, so that
	
	\begin{equation}
	\tilde{L}_a(x)=L_a (x+U(x)),
	\end{equation}	
	\begin{equation}
	\tilde{L}_t(x)=L_t (x+V(x)),
	\end{equation}
	where $\tilde{L}_a $  and $\tilde{L}_t $ denote the warped  $L_a$ and $L_t$, respectively. 
	
\begin{figure}[htb] 
		\centering
		\includegraphics[width=1\textwidth]{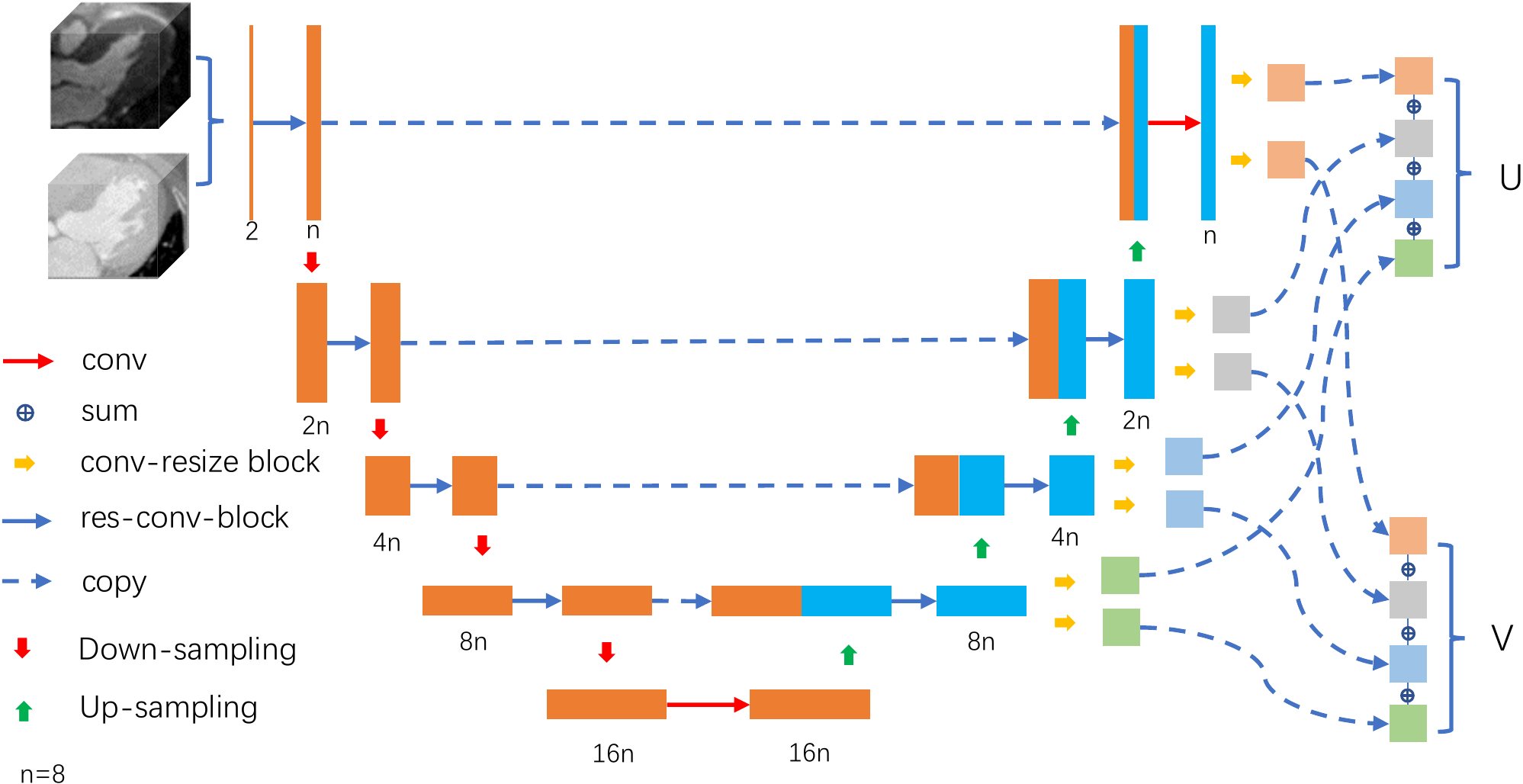}
		\caption{The architecture of the consistent registration network.}
		\label{fig:2}    
	\end{figure}

	We develop a new registration network which can jointly estimate the forward ($U$) and inverse ($V$) DDF for a pair of input images. The advantage of joint estimation is that it can reduce the ambiguous correspondence in DDFs (see next subsection). Figure \ref{fig:2} shows the overall structure of the registration network. The backbone of the network is based on the U-Shape registration model \cite{ref_article6}. Instead of using voxel-level ground-truth transformations, which is hard to obtain in cross-modality scenarios, the Dice coefficients of anatomical labels are used to train the network. Since the network is design to produce both  $U$ and $V$, pairwise registration errors caused by those two DDFs should been taken into account in the loss function. Thus, a symmetric Dice loss of the network is designed by
	\begin{equation}
	\mathcal{L}oss_{Dice}=\mathcal{D}ice(L_a,\tilde{L}_t)+\mathcal{D}ice(L_t,\tilde{L}_a) )+\lambda_1 (\Psi(U)+\Psi(V)),
	\end{equation}
	where $\lambda_1$ is the hyperparameter, $\Psi(U)$ and $\Psi(V)$ are smoothness regularizations for DDFs. 
	\subsubsection{Consistent Constraint:}
	The $\mathcal{L}oss_{Dice}$ only provides voxel-level matching criterion for transformation field estimation. It is easily trapped into a local maximum due to the ambiguous correspondence in the voxel-level DDF. Inspired by the work of Christensen et al.  \cite{ref_article15}, a consistent constraint is employed to encourage the uniqueness of the field. i.e., each voxel in $L_a$ is mapped to only one voxel in $L_t$, and vice versa. To achieve this, an invertible loss $\mathcal{L}oss_{inv}$ is engaged to force the restored warped label $L_{a}^{'}$ (or $L_{t}^{'}$) to be identical to its original label $L_a $ (or $L_t $),
	\begin{equation}
	\mathcal{L}oss_{inv}=\mathcal{D}ice(L_{a}^{'},L_a)+\mathcal{D}ice(L_{t}^{'},L_t),
	\end{equation}
	where $L_{a}^{'}(x)=\tilde{L}_a(x+V(x)))$ and $L_{t}^{'}(x)=\tilde{L}_t(x+U(x)))$. Ideally, $\mathcal{L}oss_{inv} $ is equal to $0$ when $U$ and $V$ are the inverse of each other. Therefore, it can constrain the network to produce invertible DDFs. Finally, the total trainable loss of the registration model is
	\begin{equation}
	\label{equ:5}
	\mathcal{L}oss_{reg}=\mathcal{L}oss_{Dice}+\lambda_2 \mathcal{L}oss_{inv}.
	\end{equation}
	Here, $\lambda_2$ is the hyperparameter of the model. As only anatomical labels are needed to train the network, the consistent registration network is naturally applicable to cross-modality registration. 
	
	\subsection{Similarity Network}
	\label{sec:sim}
	
	\subsubsection{Network Architecture:}
	
	Based on the registration network, $(I_a,L_a )$  can be deformed toward $I_t$ and become the warped atlas $(\tilde{I}_a,\tilde{L}_a)$, where $\tilde{L}_a$ is a candidate segmentation of $I_t$. Given $N$ atlases, the registration network will produce $N$ corresponding segmentations. Then, the target label of $I_t$ is derived by  combining the contribution of each warped atlas via PLF strategy. For a spatial point $x$, the optimal target label $\ddot{L}_t(x)$ is defined as 
	\begin{equation}
	\ddot{L}_t(x)=\mathop{\arg\max}_{l=\{l_1,l_2,…..l_k\}}  \sum \limits_{i=1}^{N}  w^i (x)\delta(\tilde{L}_a^i(x),l),
	\end{equation}
	where  $\{l_1,l_2,…..l_k\}$ is the label set, $w^i (x)$ is the contribution weight of i-th warped atlas,  and $\delta(\tilde{L}_a^i(x),l)$ is the Kronecker delta function. Usually, $w^i (x)$ is measured according to the intensity similarity among local patches. Inspired by the idea of prototypical method \cite{ref_article17}, there exists an embedding that can capture more discriminative features for similarity measurement. We design a convolution network  to map the original patches into a more distinguishable embedding space, and similarities (contribution weights) can be computed according to the distance between the embedded atlas and target patches. 
	
\begin{figure}[tb] 
		\centering
		\includegraphics[width=0.9\textwidth]{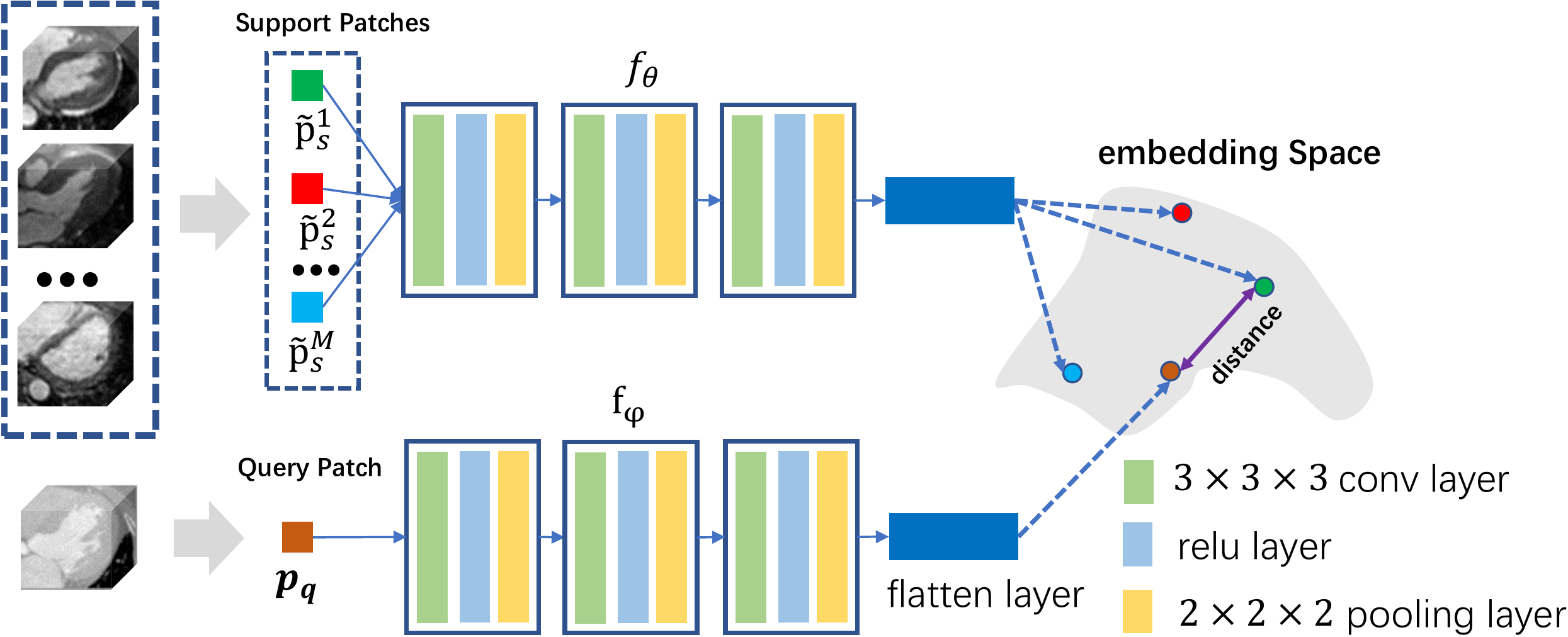}
		\caption{The architecture of the similarity network.}
		\label{fig:3}    
	\end{figure}

	Figure \ref{fig:3} shows the architecture of the similarity network. It contains two convolution ways ($f_\varphi $ and $f_\theta$ ), which can map the target and atlas patches into a embedding space separately. According to the prototypical method, we define the patch from target image as the query patch ($p_q$), and define the patches sampled from warped atlases as support patches ($p_s=\{\tilde{p}_{s}^{1},\tilde{p}_s^2,\dots,\tilde{p}_s^M\}$). The similarity $sim_i$ of $p_q$ and $\tilde{p}_s^i$ is calculated based on a softmax over the Euclidean distance of embedded atlas $f_\theta (\tilde{p}_s^i)$ and target patch $f_\varphi (p_q)$,
	\begin{equation}
	\label{equ:7}
	sim_i=\frac{exp(-d(f_\varphi(p_q),f_\theta (\tilde{p}_s^i)))}{\sum \limits_{m=1}^{M} exp(-d(f_\varphi (p_q),f_\theta(\tilde{p}_s^m)))}.
	\end{equation}
	\subsubsection{Training Algorithm:}
	We explore to train the similarity network by using the anatomical label information.  Let $y_i$ denotes the ground-truth similarity between  $p_q$ and $\tilde{p}_s^i$. The parameters of $f_\theta$ and $f_\varphi$ can be optimized by minimizing the cross-entropy loss ($J$) of the predicted and ground-truth similarities, 
	\begin{equation}
	J=-\sum \limits_{i=1}^{M} y_ilog(sim_i).
	\end{equation}
	However, $y_i$ is hard to obtain in cross-modality scenarios. To train the network, the support patches ($\tilde{p}_s^i$) which have significant shape difference or similarity to the query patch ($p_q$) are used, and their corresponding $y_i$ is decided by using the anatomical labels,
	\begin{equation}
	\label{equ:9}
	y_i =\left\{
	\begin{array}{lr}
	1 \qquad &\mathcal{D}ice(l_q, \tilde{l}_s^i)>thr_1   \\
	0 \qquad & \mathcal{D}ice(l_q, \tilde{l}_s^i) < thr_2\\
	\end{array}.
	\right.
	\end{equation}
	where $thr_1$  and $thr_2$ are hard thresholds, $l_q$ and $\tilde{l}_s^i$ denote the anatomical label of  $p_q$ and $\tilde{p}_s^i$, respectively. The network is trained in a fashion of few-shot learning,   each training sample is compose of a query patch ($p_q$) and two support patches ($ \tilde{p}^j_s ,\tilde{p}^k_s$) with significant shape differences ($y_j \ne y_k$). In this way, the convolution layers can learn to capture discriminative features for measuring similarity of cross-modality. Algorithm \ref{alg:1} provides the pseudocode. For the conciseness, the code only describe one atlas and one target setup here, while the reader can easily extend to \textit{N} atlas and \textit{K} targets in practice.
	
	\begin{algorithm}[tb]
		\DontPrintSemicolon
		\caption{Pseudocode for training the similarity network }
		\KwIn{$(\tilde{I}_a ,\tilde{L}_a)$;  $(I_t,L_t)$; the batch size $B$; the training iteration $C$}
		\KwOut{$\theta$,$\varphi$}
		\label{alg:1}
		\bf{Init $\theta$,$\varphi$}
		
		\For{\bf{$c=1$ to $C$}}
		{
			$J\leftarrow0$
			
			\For{\bf{$b=1$ to $B$}}
			{
				%				$(\tilde{p}_s^j,\tilde{l}_s^j),(\tilde{p}_s^k,\tilde{l}_s^k) ,(p_q,l_q) \leftarrow Sample((\tilde{I}_a,\tilde{L}_a),(I_a L_a))$
				$(\tilde{p}_s^j,y_j),(\tilde{p}_s^k,y_k) ,p_q \leftarrow Sample((\tilde{I}_a,\tilde{L}_a),(I_t,L_t),thr_1,thr_2)$
				
				%				$y_j,y_k \leftarrow Decide(l_q,\tilde{l}_s^j,\tilde{l}_s^k,thr_1,thr_2)$
				$sim_j,sim_k \leftarrow Calculate(\tilde{p}_s^j,\tilde{p}_s^k,p_q)$ \tcp*{see Eq.(\ref{equ:7})}
				
				$J\leftarrow J+ \frac{1}{B}(\sum_{i=\{j,k\}} y_ilog(sim_i))$  
				
			}
			
			$\theta^{new}\leftarrow \theta^{old}\-- \epsilon\bigtriangledown_\theta J $ 
			
			$\varphi^{new}\leftarrow \varphi^{old}\-- \epsilon\bigtriangledown_\varphi J$
		}
	\end{algorithm}    
	
	\section{Experiment}
	\paragraph{\textbf{Experiment setup:}}
	We evaluated the framework by myocardial segmentation of the MM-WHS dataset \cite{ref_article0}. The dataset provides 40 (20 CT and 20 MRI) images with corresponding manual segmentations of whole heart. For cross-modality setup, MR (CT) images with their labels are used as the atlases and CT (MR) images are treated as the targets. We randomly selected 24 (12 CT and 12 MR) images for training the registration network, 8 (4 CT and 4 MR) images for training the similarity network. The remaining 8 (4 CT and 4 MR) images were used as test data. Form each image, a $96 \times 96 \times 96$ sub-image around LV myocardium was cropped, and all the sub-images were normalized to zero-mean with unit-variance. In order to improve the performance, both the affine and deformable transformation were adopted for data augmentation .
	\paragraph{\textbf{For training the registration network:}} In each training iteration, a pair of CT-MR intensity images is fed into the registration network (see Figure \ref{fig:2}). Then the network  produce $U$ and $V$, with which the MR and CT label can be warped to each other. By setting the hyperparameter $\lambda_1$ and $\lambda_2$ to 0.3 and 0.2, the total trainable loss (see Eq.(\ref{equ:5})) of the network can be calculated. Finally, Adam optimizer is employed to train the parameters of network. 
	\paragraph{\textbf{For training the similarity network:}} For training the network, we extracted patches along the boundary of LV myocardium (which usually cover different anatomical structure).  In each training iteration, the size of patch is set to $15 \times 15 \times 15$ voxels, while the $thr_1$  and $thr_2$ are set to 0.9 and 0.5, respectively  (see Eq.(\ref{equ:9})). Training sample $(\tilde{p}_s^j,\tilde{p}_s^k,p_q)$ is randomly selected and then mapped into the embedding space. Finally, the loss can be accumulated and backpropagated to optimize the parameters of $f_\varphi $ and $f_\theta$  (see Algorithm \ref{alg:1}). 
	
	\begin{table*}[htb] 
		\begin{floatrow}
			\capbtabbox[0.45\textwidth]{
				\begin{tabular}{cc}
					\hline
					Method &  Dice (Myo)\\
					\hline
					Demons CT-MR \cite{thirion1998image}& 	36.1 $\pm$ 10.9  	\% \\
					Demons MR-CT \cite{thirion1998image}& 	36.9 $\pm$ 12.8 	\% \\
					SyNOnly CT-MR \cite{avants2009advanced} & 	52.6 $\pm$ 13.9 	\% \\
					SyNOnly MR-CT \cite{avants2009advanced} & 	55.3 $\pm$  10.8	\% \\
					CT-MR 	\cite{ref_article6} & 	70.5 $\pm$ 4.8  	\% \\
					MR-CT \cite{ref_article6} & 	73.4 $\pm$ 4.7 \%  \\
					\hline
					Our	CT-MR  & 	\textbf{74.4 $\pm$ 5.2} \% 	 \\
					Our	MR-CT  & 	\textbf{76.4 $\pm$ 4.7} \%  \\
					\hline
				\end{tabular}
			}{
				\caption{Comparison between the proposed registration network and other state-of-the-art methods.}
				\label{tab:1}
			}
			\capbtabbox[0.45\textwidth]{
				\begin{tabular}{cc}
					\hline
					Method &  Dice (Myo)\\
					\hline
					U-Net \cite{ref_article19} & 86.1 $\pm$ 4.2  \% \\
					Seg-CNN \cite{ref_article16} &   \textbf{ 87.2 $\pm$ 3.9 } \% \\
					MV MR-CT &	 84.4 $\pm$ 3.6 \%  \\
					NLWV MR-CT & 84.9 $\pm$ 4.0 \% \\
					Our  MR-CT & 84.7 $\pm$ 3.9 	\%  \\
					\hline
					U-Net \cite{ref_article19} & 68.1  $\pm$ 25.3 \% \\
					Seg-CNN \cite{ref_article16} &  75.2  $\pm$ 12.1 \% \\
					MV CT-MR &	 80.8 $\pm$ 4.8 \%  \\
					NLWV CT-MR & 81.6 $\pm$ 4.7 \%  \\
					Our  CT-MR & \textbf{81.7 $\pm$ 4.7 }	\%  \\
					
					\hline
				\end{tabular}
			}{
				\caption{Comparison between the proposed MAS and other state-of-the-art  methods.}
				\label{tab:2}
			}
		\end{floatrow}
	\end{table*}
	
	\paragraph{\textbf{Results:}}
	The performance of the registration network is evaluated by using the Dice score between the warped atlas label and the target gold standard label. Table \ref{tab:1} shows the average Dice scores over $48$ (12 CT $\times$ 4 MR or 12 MR $\times$ 4 CT) LV myocardium registrations. CT-MR (MR-CT) indicates when CT (MR) images are used as atlas and  MR (CT) images are used as target. Compared to the U-shape registration network \cite{ref_article6}, the proposed network achieves  almost 3.5\% improvement of Dice score. Additionally, our method outperforms the conventional methods (SyNOnly \cite{avants2009advanced} and Demons \cite{thirion1998image}). This is reasonable as our method takes advantage of the high-level information (anatomical label) to train the registration model, which makes it more suitable for the challenging dataset of MM-WHS.
	
\begin{figure}[htb] 
		\centering
		\includegraphics[width=0.9\textwidth]{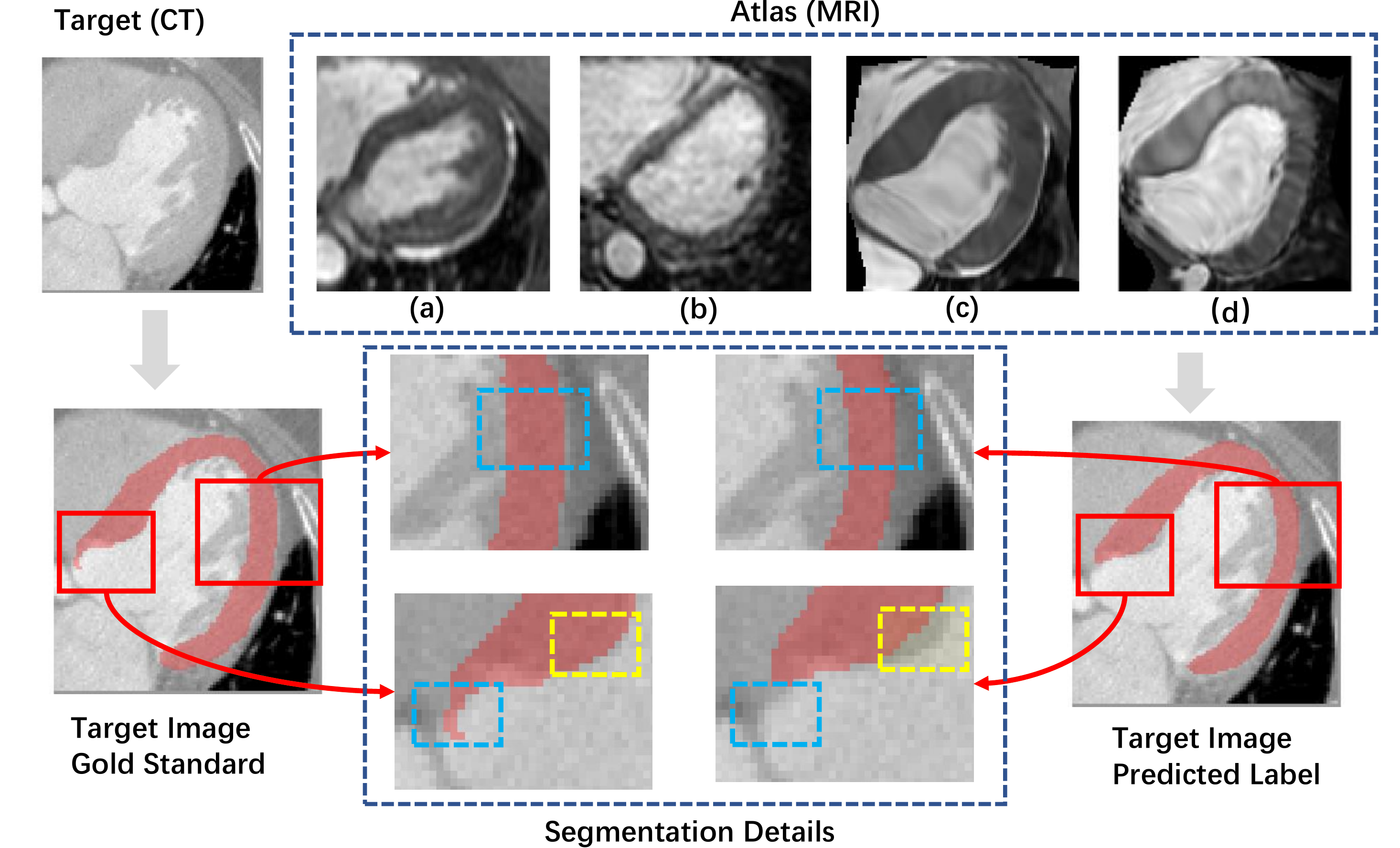}
		\caption{Visualization of the proposed framework. Image of (a) and (b) are atlas images, (c) and (d) are corresponding warp atlas images. All the images are from axial-view. The segmentation details show different slices, where the region in the yellow box shows the error by our method, while the region in the blue boxes indicate that the proposed method performs not worse than the golden standard. (The reader is referred to the colourful web version of this article)}
		\label{fig:4}    
	\end{figure}

	Table \ref{tab:2} shows the result of three different MAS methods based on our registration network. ie, MV, non-local weighted voting (NLWV) \cite{ref_article18} and the proposed framework. Compared to other state-of-the-art methods \cite{ref_article16,ref_article19}, our framework can achieve promising performance in cross-modality myocardial segmentation. Especially in MR images, compared to the Seg-CNN \cite{ref_article16} who won the first place of MM-WHS Challenge, our framework improves the Dice score by almost 6\%. However, our MR-CT result, which is set up to use CT atlases to segment an MR target, is worse than other state-of-the-art methods. This is because the quality of atlas will affect MAS performance. Generally, MR is considered more challenging data (lower quality) compared to CT \cite{ref_article0}. The use of low-quality MR atlases limits the segmentation accuracy of our MR-CT. Thus, the Seg-CNN \cite{ref_article16}, which is trained on purely CT data, can obtain almost 3\% better Dice score than our MR-CT method. In addition, Figure \ref{fig:4} demonstrates a series of intermediate results and segmentation details.
	
\begin{figure}[htb] 
		\centering
		\includegraphics[width=0.9\textwidth]{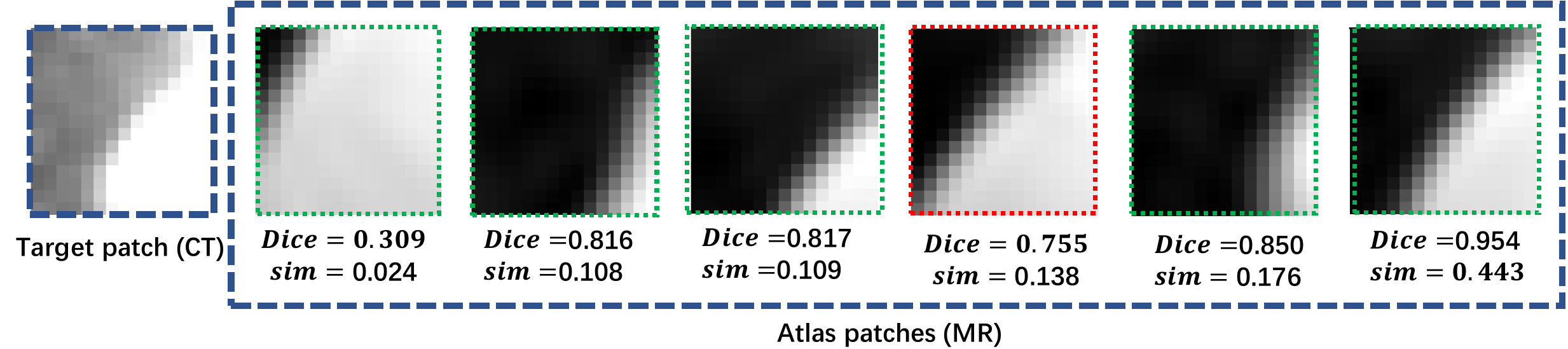}
		\caption{Visualization of estimated similarities from the proposed network. The Dice scores ($Dice$) are calculated by the gold standard label, while the similarities ($sim$) are estimated by feeding intensity patches into the similarity network. Please note that the $sim$ are normalized by the softmax function of similarity network (see Eq.(\ref{equ:7})). Ideally, the $sim$ should be positively related to the $Dice$. This figure shows both the correct (green box) and failed (red box) cases of our similarity estimation method.}
		\label{fig:5}    
	\end{figure}

	Figure \ref{fig:5} visualizes the performance of similarity network. The target patch is randomly selected from CT image, and the atlas patches are randomly cropped from MR images. Since Dice coefficient computes similarity of patches by using golden standard labels, it can be considered as the golden standard for cross-modality similarity estimation. Results show that the estimated similarities are well correlated to the Dice coefficient.
	
	\section{Conclusion}
	We have proposed a cross-modality MAS framework to segment a target image using the atlas from another modality. Also, we have described the consistent registration and similarity estimation algorithm based on DNN models. The experiment demonstrates that the proposed framework is capable of segmenting myocardium from CT or MR images. Future research aims to extend the framework to other substructure of the whole heart, and investigate the performance on different datasets.
	
	\bibliographystyle{splncs04}
	
	\bibliography{ref}

\end{document}